\documentclass{amia_sy}
\usepackage[margin=1in]{geometry}
\usepackage{color,comment,ulem,soul}
\usepackage{xcolor}

\usepackage{color}
\usepackage[utf8]{inputenc} 
\usepackage[T1]{fontenc}    
\usepackage{natbib}
\usepackage[colorlinks,citecolor=blue,linkcolor=blue,urlcolor=blue]{hyperref}
\usepackage{url}            
\usepackage{booktabs}       
\usepackage{amsfonts}       
\usepackage{nicefrac}       
\usepackage{microtype}      
\usepackage{xcolor}
\usepackage{latexsym, epsfig, amssymb, amsmath, amsthm, mathrsfs,mathtools,graphicx}

\usepackage{enumerate}
\usepackage{multirow}
\usepackage{multicol}
\usepackage{array}
\usepackage{enumitem}
\usepackage{epstopdf}
\usepackage{caption}
\usepackage{subcaption}
\usepackage{wrapfig}
\usepackage{adjustbox}
\usepackage{algorithmic,algorithm,bm}
\usepackage{tikz}
\usepackage[most]{tcolorbox}
\usepackage[toc,page,header]{appendix}
\usepackage{minitoc}
\usepackage[capitalize,noabbrev]{cleveref}
\usepackage{tablefootnote}
\usepackage[superscript, nomove]{cite}
\usepackage{caption}
\usepackage{makecell} 

\usepackage{color,comment,ulem,soul}

\newcommand{\mc}{\mathcal}
\newcommand{\m}[1]{{\bf{#1}}}

\newcommand{\g}[1]{\mbox{\boldmath $#1$}}
\newcommand{\mb}[1]{{\mathbb{#1}}}

\title{Clustering Alzheimer’s Disease Subtypes via Similarity Learning and Graph Diffusion}

\author{Tianyi Wei$^{1,}$\footnote{Equal contribution by T. Wei, S. Yang and D. Ataee Tarzanagh.}, Shu Yang$^{1,\textmd{I}}$, Davoud Ataee Tarzanagh$^{1,\textmd{I}}$, Jingxuan Bao$^1$, Jia Xu$^1$, Patryk Orzechowski$^{1,2}$, Joost B. Wagenaar$^1$, Qi Long$^1$, Li Shen$^{1,}$\footnote{Correspondence to li.shen@pennmedicine.upenn.edu.}, for the Alzheimer's Disease Neuroimaging Initiative\footnote{Data used in preparation of this article were obtained from the Alzheimer's Disease Neuroimaging Initiative (ADNI) database (adni.loni.usc.edu). As such, the investigators within the ADNI contributed to the design and implementation of ADNI and/or provided data but did not participate in analysis or writing of this report. A complete listing of ADNI investigators can be found at: http://adni.loni.usc.edu/wp-content/uploads/how\textunderscore to\textunderscore apply/ADNI\textunderscore Acknowledgement\textunderscore List.pdf}}

\institutes{
$^{1}$ Department of Biostatistics, Epidemiology and Informatics, Perelman School of Medicine, University of Pennsylvania, Philadelphia, PA, USA.\\
$^{2}$ Department of Automatics and Robotics, AGH University of Science and Technology, al. Mickiewicza 30, 30-059 Krakow, Poland \\
}

\begin{document}

\maketitle

\begin{abstract}
Alzheimer's disease (AD) is a complex neurodegenerative disorder that affects millions of people worldwide. Due to the heterogeneous nature of AD, its diagnosis and treatment pose critical challenges. Consequently, there is a growing research interest in identifying homogeneous AD subtypes that can assist in addressing these challenges in recent years. In this study, we aim to identify subtypes of AD that represent distinctive clinical features and underlying pathology by utilizing unsupervised clustering with graph diffusion and similarity learning. We adopted SIMLR, a multi-kernel similarity learning framework, and graph diffusion to perform clustering on a group of 829 patients with AD and mild cognitive impairment (MCI, a prodromal stage of AD) based on their cortical thickness measurements extracted from magnetic resonance imaging (MRI) scans. Although the clustering approach we utilized has not been explored for the task of AD subtyping before, it demonstrated significantly better performance than several commonly used clustering methods. Specifically, we showed the power of graph diffusion in reducing the effects of noise in the subtype detection. Our results revealed five subtypes that differed remarkably in their biomarkers, cognitive status, and some other clinical features. To evaluate the resultant subtypes further, a genetic association study was carried out and successfully identified potential genetic underpinnings of different AD subtypes. 
Our source code is available at: \url{https://github.com/PennShenLab/AD-SIMLR}.
\vspace{.25cm}
\end{abstract}

\section{INTRODUCTION}

Alzheimer's disease (AD) is a neurodegenerative disorder characterized by progressive cognitive decline and memory loss. Among over 55 million people affected by dementia worldwide in 2023, AD is the single most common form and accounts for more than two-thirds of all dementia cases \footnote{https://www.who.int/news-room/fact-sheets/detail/dementia}. The projected number of individuals impacted by AD and AD related dementias (ADRD) could surge to more than 100 million by 2050, if there is no significant progress in drug discovery \cite{PRINCE201363, 2022e105}. However, despite its prevalence, the pathological heterogeneity of AD constitutes a substantial barrier to deciphering the etiological mechanism of the disease and developing corresponding treatments \cite{Ferreira436, MURRAY2011785,bao2021identifying,bao2021identifying2,bao2023integrative}. Therefore, as a result, the identification of AD subtypes or clusters is becoming increasingly important as it can help us understand the different neurodegenerative paths and provide more nuanced information to assist in drug development efforts \cite{HABES202070, 10.1093/braincomms/fcaa192,wen2022characterizing}.  

Recently, the success of machine learning (ML) in achieving better performance on identifying AD subtypes has relied in significant part on cluster analysis, in conjunction with introducing more effective ML methods. This success is partially driven by the availability of a wide range of clustering methods such as spectral clustering.  
Existing similarity-based methods typically employ similarity/distance measures between individuals and then apply unsupervised \cite{doi:10.1073/pnas.1611073113, https://doi.org/10.1002/ana.25142, HWANG201658, Park2017} or semi-supervised \cite{Chimera7293208,Yang2021} ML methods, among others, to cluster patients with AD into different subtypes. They obtain similarity  measures directly with predefined metrics like 1/2-norm distance \cite{https://doi.org/10.1002/ana.25142, HWANG201658} and Pearson correlation \cite{Park2017, 10.1117/12.2511573} or indirectly from methods like RBF kernels \cite{Chimera7293208} based on medical imaging or other clinical/biological features of each subject. A key problem with the aforementioned methods is that they rely on similarity metrics which may suffer from high noise level in the data and may not generalize well across platforms/biological experiments. 

It is known that if the goal is structured graph (similarity) learning, community inference and graph estimation should be done \textit{jointly}. In fact, performing structure inference after graph estimation may result in a sub-optimal procedure \cite{marlin2009sparse}. To overcome this issue, some of the initial work focused on either inferring connectivity information or performing graph estimation in case the connectivity or community information is known~\textit{a priori} \cite{Danaher13,guo2011joint,gan2019bayesian,ma2016joint,lee2015joint}. Recent developments consider the two tasks jointly and estimate the structured graphs arising from heterogeneous observations~\cite{kumar2020unified,hosseini2016learning,hao2018simultaneous, tarzanagh2018estimation,kumar2019structured, gheche2020multilayer,eisenach2020high,tarzanagh2021fair,pircalabelu2020community,wang2017visualization}. In order to tackle the issues of the generalization of similarity metrics and structured graph estimation, a manifold learning framework called SIMLR~\footnote{Single-cell interpretation via multikernel learning (SIMLR)}~\cite{wang2017visualization} was developed previously that can learn an appropriate similarity metric from the input data. Briefly, SIMLR learns a similarity matrix that best fits the structure of the data by combining multiple kernels and jointly estimates an undirected graph and communities of nodes. Although initially developed for single-cell RNA-seq analysis, SIMLR has later been successfully applied to various biological and medical studies \cite{kausar2019density, qi2020clustering, deng2019scalable, vega2022simlr}. By training a model to directly learn the similarity between individuals based on their features, SIMLR is able to identify subtle patterns and relationships between individuals that may be missed by traditional clustering methods. This ability to capture complex relationships is particularly important in the context of AD, where heterogeneity and diverse clinical presentations are prevalent. 

While similarity/graph learning algorithms can improve the performance of conventional methods such as $K$-means on a wide variety of clustering tasks, they may suffer from high levels of noise. To overcome this issue, previous works used a graph diffusion strategy that can enhance the weak similarities of the nodes by reducing the effect of the noise; albeit in the signal processing ~\cite{yang2010modeling} and genomics data analysis \cite{wang2014similarity}. On the other hand, the effect of graph diffusion on AD dataset and brain MRI analysis has been elusive perhaps due to the additional challenges surrounding similarity learning. 

In addition, traditional diagnostic criteria for AD have focused on the clinical syndrome, which typically presented as a predominantly amnestic phenotype, as well as a pre-dementia phase referred to as Mild Cognitive Impairment (MCI, a prodromal stage of AD); more recently, there have been growing efforts to biologically characterize AD based on the presence of biomarkers such as brain imaging traits like cortical atrophy measured from magnetic resonance imaging (MRI), amyloid or tau deposition from positron emission tomography (PET), genetic variants from genotyping, etc. One such example is the Alzheimer's Disease Neuroimaging Initiative (ADNI) \cite{saykin2015aad,shen2014bior,weiner2017recent}. ADNI was launched in 2003 with the primary goal to measure the progression of MCI and early AD, and it consists of comprehensive clinical, cognitive, and imaging data from a large cohort of individuals across the AD spectrum. ADNI provides a rich resource for investigating the heterogeneity of AD, and it has been widely used to study AD biomarkers \cite{shen2011mbia,yan2014bioinfo,wan2014tmi} as well as to identify AD subtypes and evaluate the clinical relevance of these subtypes \cite{feng2022bmcbioinfo,feng2020bibe}.

In this study, we focus on the cortical thickness measurements from ADNI which have been reported to reflect pathophysiological and clinical changes in AD \cite{IMA:IMA2, HWANG201658, Park2017}, and we propose to leverage the power of similarity learning and graph diffusion to identify AD subtypes by analyzing data from three distinct populations: cognitively normal (CN) subjects, individuals with MCI, and AD patients. Focusing on these populations allows us to better understand the disease's progression and distinguish between different subtypes, thereby providing valuable insights into the complex nature of AD. We perform graph-based similarity learning and clustering with spectral clustering, SIMLR,  and their diffusion variants to subtype subjects derived from ADNI database, and we compare the performance with alternative clustering methods to demonstrate the effectiveness of our proposed approach. Specifically, we show that graph diffusion enhances similarity learning in brain MRI and reduces the influence of noise in popular graph-based approaches such as spectral clustering. Moreover, to validate our clustering results, we also conducted a targeted genetic association analysis with our subtypes and successfully identified potential genetic basis of the different AD subtypes which display different risks.

\section{MATERIALS AND METHODS} 

\begin{figure}[t]
\centering
\includegraphics[width=0.9\textwidth]{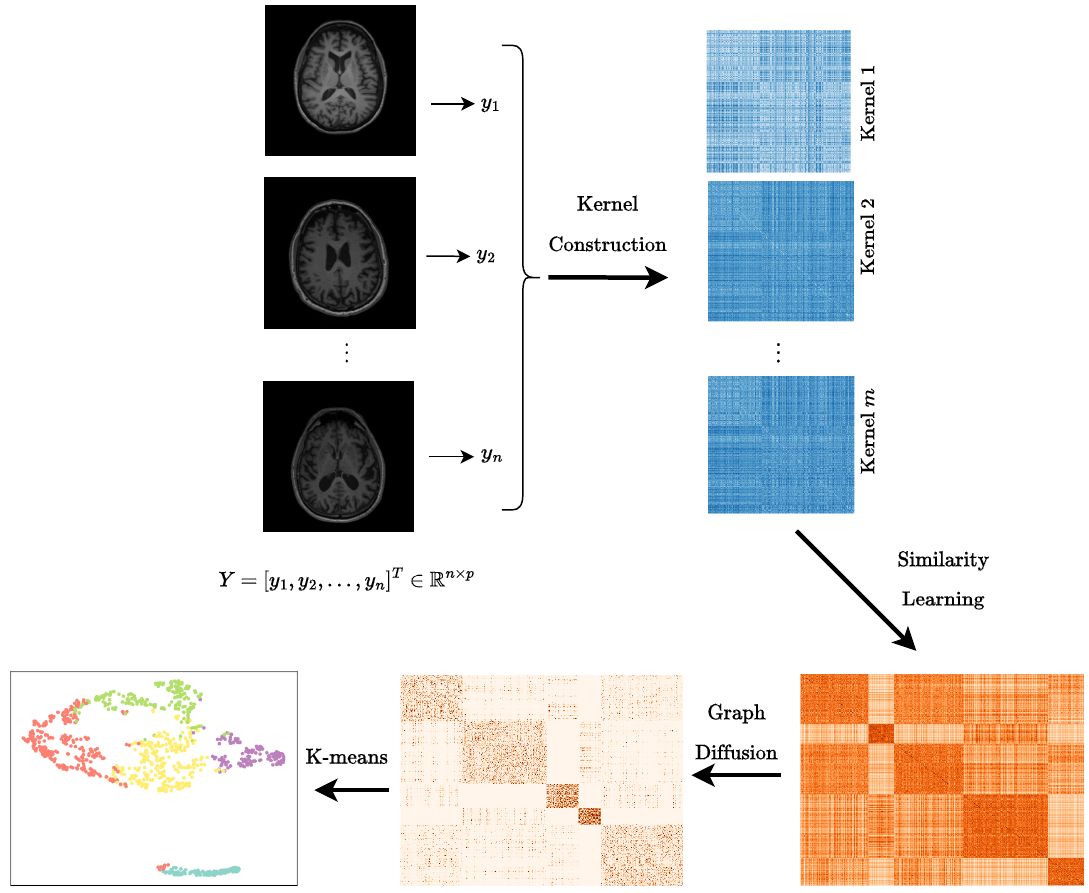}
\caption{Schematic overview of the proposed methods (spectral clustering, SIMLR, and their diffusion variants) for AD subtype detection. For SIMLR+graph diffusion, one needs to repeat the similarity learning and graph diffusion steps for a few iterations.  }
\label{fig:overview}
\end{figure}

Figure~\ref{fig:overview} shows the overall design of this work, where we have applied several clustering methods (i.e., spectral clustering, SIMIR, and their diffusion variants) to the ADNI MRI-based measures for AD subtype detection. Below we discuss the materials used and methods proposed in this study.

\subsection{Data Preparation} \label{EXP:data}

Data used in the preparation of this article were obtained from the ADNI database (adni.loni.usc.edu) \cite{saykin2015aad,shen2014bior,weiner2017recent}. The ADNI was launched in 2003 as a public-private partnership, led by Principal Investigator Michael W. Weiner, MD. The primary goal of ADNI has been to test whether serial MRI, PET, other biological markers, and clinical and neuropsychological assessment can be combined to measure the progression of MCI and early AD. Up-to-date information about the ADNI is available at \url{www.adni-info.org}.

Specifically, we used the imaging and demographic data from the Alzheimer's Disease Prediction Of Longitudinal Evolution (TADPOLE) Challenge \cite{Marinescu2019-rw} datasets that were derived from ADNI, which consists of MRI-based cortical thickness measures and demographic data. We selected the baseline visits of 1,179 subjects (350 cognitively normal (CN) subjects, 631 MCI subjects, 198 AD subjects) from the D1 and D2 data sets of TADPOLE challenge along with the subjects' 72 average cortical thickness measurements. Among these 72 measurements, 9 measurements were excluded because the data were missing for more than 50\% of the subjects, which resulted in a total of 63 measurements for each subject. All cortical thickness measurements that are more than 1.5 interquartile range below the first quantile (lower threshold) or more than 1.5 interquartile range above the third quantile (upper threshold) are scaled up or down to the lower or upper threshold accordingly. The 350 CN subjects were further reduced to 199 subjects, forming an age- and gender- matched control group for comparison with the 631 MCI subjects and 198 AD subjects. The final study cohort consists of 1,028 subjects including 199 CN subjects, 631 MCI subjects, and 198 AD subjects. 

We also downloaded the genotyping data of the studied subjects. For these data, we performed quality control (QC) using the following criteria: genotyping call rate $>$ 95\%, minor allele frequency $>$ 5\%, and Hardy-Weinberg  Equilibrium $>$ 1e-6. To evaluate our clustering findings, we prioritized 57 AD-related single nucleotide variants (SNVs) based on the published large-scale genome-wide association studies (GWAS) of AD and closely related phenotypes \cite{king2019drug,bellenguez2022new,kunkle2019genetic,novikova2021integration,mountjoy2021open,amlie2018inferno} that were evaluated by the gene verification committee of the Alzheimer's disease sequencing project (ADSP) \cite{beecham2017alzheimer}.

\subsection{Clustering and Graph Diffusion Methods}
Let $\m{Y}$ be an $n\times p$ data matrix, with rows $\m{y}_{1},\ldots,\m{y}_{n}$ indicating $n$ subjects. We associate to each row in $\m{Y}$ a node in a graph $\mc{G}=( \mc{V}, \mc{E})$, where $\mc{V}= \{1,2,\ldots,n \}$ is the vertex set and $\mc{E}\in \mc{V}\times\mc{V}$ is the edge set.  We consider a simple undirected graph, without self-loops, and whose edge set contains only distinct pairs. Graphs are conveniently represented by a $n \times n$ matrix, denoted by $\g{\Theta}=(\theta_{ij})$, whose nonzero entries correspond to edges in the graph. The precise definition of this usually depends on modeling assumptions,  properties of the desired graph, and application domain.  Throughout, we assume further the existence of $K$ communities of nodes and denote by $\mc{C}_k$ the subset of nodes from $\mc{G}$ that belongs to the $k$th community.

\paragraph{Spectral Clustering.}   Given $K\in \mb{N}$, \textit{unnormalized} spectral clustering (SC) aims to partition  $\mc{V}$ into $K$ clusters with minimum value of the \texttt{RatioCut} objective function as follows \cite{von2007tutorial}:
For a clustering $\mc{V}=\mc{C}_1{\cup}\ldots {\cup}\mc{C}_K$, we have
\begin{equation}
    \begin{split}
       \texttt{RatioCut}(\mc{C}_1,\ldots,\mc{C}_K)&:=\sum_{l=1}^K \frac{\text{cut}(\mc{C}_l, \mc{V}\setminus \mc{C}_l)}{|\mc{C}_l|}, \\
   \text{where}~~~~&\text{cut}(\mc{C}_l,V\setminus \mc{C}_l):=\sum_{i\in \mc{C}_l, j\in \mc{V}\setminus \mc{C}_l}\theta_{ij}.
   \label{def_ratio_cut}
    \end{split}
       \tag{RatioCut}
\end{equation}
\eqref{def_ratio_cut} aims to divide graph nodes $\mc{V}$ into  disjoint subsets while minimizing the ratio of the number of edges between the two subsets to the total number of edges in the graph.  
Let $\m{D}$ be the degree matrix, that is a diagonal matrix with the vertex degrees $d_i=\sum_{j\in[n]}\theta_{ij}$, $i\in [n]$, on the diagonal. 
Let $\m{L}=\m{D}-\g{\Theta}$ denote the unnormalized graph Laplacian matrix.  
For each candidate partition of $n$ nodes into $K$ communities, we associate it with a \textit{partition matrix} $\m{V} \in [0, 1]^{n \times k}$, such that $v_{ik}=1/\sqrt{|\mc{C}_k|}$ if and only if nodes $i$ and $k$ are assigned to the same community. Then, $\texttt{RatioCut}(\mc{C}_1,\ldots,\mc{C}_k)=\text{trace}(\m{V}^\top\m{LV})$. 

Hence, in order to minimize the \eqref{def_ratio_cut} function over all possible clusterings, we could minimize $\text{trace}(\m{V}^\top \m{L} \m{V})$ over all partition matrices. 
Spectral clustering relaxes this minimization problem by replacing the partition constraint on $\m{V}$ with $\m{V}^\top \m{V}=\m{I}_K$, that is it solves
\begin{align}\label{eqn:obj:sc}
 \tag{SC}
 \underset{\m{V}\in\mb{R}^{n\times k}}{\text{minimize}}~~\text{trace}(\m{V}^\top (\m{D}-\g{\Theta}) \m{V}) ~~ \text{subject to }~~\m{V}^\top \m{V}=\m{I}_K.
\end{align}
It is well known that a solution to
\eqref{eqn:obj:sc} is given by a matrix $\m{V}$ that contains some orthonormal eigenvectors corresponding to the $K$ smallest eigenvalues of $\m{L}$ as columns \cite{lutkepohl1997handbook}. Hence, in the first step of SC, we need to compute such an optimal~$\m{V}$ by computing the $K$ smallest eigenvalues and corresponding eigenvectors.  In the second step, we infer a clustering from~$\m{V}$ by applying $K$-means to the rows of $\m{V}$.

\paragraph{SIMLR.} Given $m$ kernels $\{K_{l}\}_{l=1}^m$, SIMLR \cite{wang2017visualization} have considered the following graph estimation problem:
\begin{align}~\label{eqn:obj:simlr}
\begin{array}{ll}
\underset{\g{\Theta},\m{V},\m{w}}{\text{minimize}} &
\begin{array}{c}
 - \sum_{l,i,j} w_l K_{l} (\m{y}_i,\m{y}_j )\theta_{ij}+ \rho_1\textnormal{trace}( \m{V}^\top (\m{I}_n-\g{\Theta}) \m{V}))+ \rho_2\|\g{\Theta}\|^2+  \rho_{3} \sum_{l} w_l \log w_l  
\end{array}\\
\text{s. t.} & \begin{array}[t]{l}
 \m{V}^\top\m{V}=\m{I}_K,~~\sum_{l} {w}_l=1,~\forall l \in [m],~{w}_l\geq 0,~\sum_{j} \theta_{ij}=1,~~\forall i,j \in [n],~\theta_{ij} \geq0.
\end{array}
\end{array}
\tag{SIMLR}
\end{align}
We detail the purpose of each component of the \eqref{eqn:obj:simlr}:
\begin{itemize}
    \item $L (\g{\Theta}, \m{w};\m{Y}):= - \sum_{l,i,j} w_l K_{l} (\m{y}_i,\m{y}_j )\theta_{ij}$ is the dissimilarity function, where each linear weight value $w_l$ gives the importance of each individual kernel $K_{l} (\m{y}_i,\m{y}_j )$, which is a function of the expression of  $\m{y}_i$ and $\m{y}_j$. The intuition behind   $L (\g{\Theta}, \m{w};\m{Y})$ is that the learned similarity $\theta_{ij}$ between two random variables $i$ and $j$ should be small if the distance between them is large.
    \item The second term links the graph information encoded by the matrix $\g{\Theta}$ with the clustering matrix $\m{V}\m{V}^\top$ through a ‘trace’ operator similar to \eqref{eqn:obj:sc}. In order to balance the contributions from  graph learning  and  clustering an extra parameter $\rho_2$ is introduced. Note that the matrix $(\m{I}_n-\g{\Theta}) $ is an \textit{approximate}  graph Laplacian, and the constrained trace-minimization problem enforces approximately $K$ connected components in  $\g{\Theta}$ that consists of nodes representing the random variables and edge weights corresponding to pairwise similarity values in $\g{\Theta}$.
 \item The third term is a classical ridge penalty that
    controls the contribution of the minor nodes of $\g{\Theta}$. It prevents the learned similarities/graph from becoming too close to an identity matrix. 
    \item The last term together with the constraints on $\m{w}$ encourages avoiding the selection of a single kernel. This can generally improve the quality of learned graph matrix $\g{\Theta}$ \cite{wang2017visualization}.
\end{itemize}

SIMLR forms  $m \times n \times n$ affinity tensor and computes eigenvectors of $n \times n$ matrix in each iteration of the optimization algorithm, operations that have a computational
complexity of $O(mn^3)$ in general. For applications with $n$ on the order of thousands, SIMLR begins to become
infeasible, and problems with $n$ in the millions are entirely out of reach. To overcome the complexity of forming affinity tensor, one can  use $K$-nearest-neighbor (KNN) similarity to approximate \footnote{https://github.com/spotify/annoy} full pairwise similarity, and then only update
similarities in these pre-selected top $K$ neighbors for each node. To facilitate the implementation of $\m{V}$, one may apply an iterative approximation method in each iteration where the full eigen decomposition is replaced by a sketching one such as Nystr\"{o}m low-rank approximation \cite{fowlkes2004spectral}.  Although this type of method may require large column sampling, given the fact that SIMLR implementation can be viewed as a preprocessing before $K$-means clustering, such approximation might be helpful when $K$ is small.

\paragraph{Graph Similarity Enhancement by Diffusion.}  \label{EXP:diffusion}
Our experiments showed that both SIMLR and SC suffer from high levels of noise.  Following the discussion in \cite{wang2017visualization,wang2014similarity}, we adapted the graph diffusion \cite{yang2010modeling} strategy to enhance the similarity matrix $\g{\Theta}$ and reduce the influences of noise in both SIMLR and SC. 
Given $\g{\Theta}$,  we set   $p_{ij}= \theta_{ij} \m{1}_{j \in \mc{A}_{N}(i)}/\left(\sum_{l}  \theta_{il}\m{1}_{ l \in \mc{A}_{N}(i)}\right)$, where $\mc{A}_{N}(i)$ represents the set of indices of the nodes that are in the top $N$ neighbors of node $i$ under learned distance metric. The resulting matrix $\m{P}=(p_{ij})$ is sparse  and preserves
most of the similarity structures. Then, we iterated to obtain 
\begin{equation}
    s_{ij}^{(t+1)}= \tau   s_{ij}^{(t)} \m{P} +(1-\tau)\m{I}_n, ~~~\text{where}~~~ s_{ij}^{(0)}=\theta_{ij}, ~~t \in \{0,\ldots, T-1\}.
\end{equation}
The final iteration $\m{S}^T$ was used as the new similarity matrix $\g{\Theta}$.

Note that the optimization problem \eqref{eqn:obj:simlr} is \textit{nonconvex} and the graph model can be stuck in a saddle point \cite{tarzanagh2021fair}. However, our experiments indicated that  SIMLR converges to a structured graph within a few iterations. To evaluate the convergence, we used the eigengap defined by the difference between the $(k + 1)$-th and $k$th eigenvalues.  Note also that  $K$ is a hyperparameter to the SC and SIMLR. We used the self-tunning approach \cite{zelnik2004self} to select cluster size as follows: consider a data set $\m{Y}$ with $K$ clusters, we aim to find an indication matrix $\m{U} (\m{Q}):= \m{Y} \m{Q}$, where $\m{Y} \in \mb{R}^{n\times K}$ is the matrix of the top eigenvectors of the Laplacian, and $\m{Q} \in \mb{R}^{K\times K}$ is a rotation matrix. We search for $\m{Q}$ such that it minimizes the \textit{separation cost}, i.e.,  $f(\m{Q}):=\sum_{ij} [\m{U} (\m{Q})]_{i,j}^2/\left(\max_j [\m{U} (\m{Q})]_{i,j}\right)^2$.  Minimizing $f$ over all possible $\m{Q}$ will provide the best alignment with the canonical coordinate system. We search for the number of clusters that result in the largest drop in $f(\m{Q})$; for further details, please see \cite{zelnik2004self,wang2017visualization}.

The low-rank regularization parameter $\rho_1$ was estimated by the distance gap between $k + 1$ and $k$-th neighbors in a data-driven fashion. The ridge penalty parameter $\rho_2$ was set to be equal to $\rho_1$. The hyper-parameter $\rho_3$ was set to one. The number of neighbors was set to $N=10$. The diffusion parameter $\tau$ was set to $0.8$. All other hyper-parameters used in SIMLR followed from the original paper\cite{wang2017visualization}.

\section{RESULTS} \label{Results}

\subsection{Cohort Description}  
A summary of the cohort used for this study is shown in Table \ref{tab:dataset}. As described earlier in Section \ref{EXP:data}, the cohort included 1,028 individuals with 199 CN subjects, 631 MCI subjects, and 198 AD subjects. The CN subjects were matched on age and gender to serve as a control group for comparison with the MCI and AD groups. For each subject, a total of 63 cortical thickness measurements were extracted from brain imaging data in the analysis as features, along with other clinical traits. All 829 subjects (631 MCI subjects, 198 AD subjects) were clustered into five subtypes. The subtypes were compared across various demographic characteristics, cognitive test scores, and biomarkers from the TADPOLE datasets. These features include: age, years of education, Alzheimer's Disease Assessment Scale 13 (ADAS13), sum of boxes of the clinical dementia rating (CDRSB), Rey Auditory Verbal Learning Test (RALVT) score, Mini-Mental State Examination (MMSE) score, Functional Activities Questionnaire (FAQ) score, fluorodeoxyglucose (FDG), Florbetapir (AV45), amyloid beta (ABETA), tau protein, and phosphorylated tau (pTau). Among the biomarkers, the amyloid beta and the tau protein are two hallmarks of AD. In the TADPOLE dataset, amyloid beta levels were measured both in cerebrospinal fluid (CSF) (represented by ABETA) and the brain (represented by AV45) where lower level of CSF amyloid beta indicates higher level of amyloid beta in the brain. Tau protein levels were also measured in CSF. Tau protein contributes to the aggregation of neurofibrillary tangles which is a key pathological feature to AD. Elevated levels of tau protein in the CSF are associated with increased risk in AD. FDG is a radioactive tracer used in positron emission tomography (PET) imaging to measure glucose metabolism in the brain. 

\begin{table}[tb]
\centering
\caption{Biomarkers and clinical measurements of 1,028 patients in the study. Mean $\pm$ Standard deviation were reported, when applicable. P-values were computed among cognitively normal (CN), mild cognitive impairment (MCI) and Alzheimer's Disease (AD) groups, based on one-way ANOVA.
}
\resizebox{\columnwidth}{!}{
\begin{tabular}{l c c c c c}
\hline
& \textbf{ALL} $(n=1,028)$ & \textbf{CN} $(n=199)$ & \textbf{MCI} $(n=631)$ & \textbf{AD} $(n=198)$ & \textbf{P-values}\\
\hline
FEMALE & 456 & 88 & 274 & 94 & \\
AGE & 72.89 $\pm$ 7.52 & 72.30 $\pm$ 7.41 & 72.67 $\pm$ 7.44 & 74.22 $\pm$ 7.70 & 0.047 \\
PTEDUCAT & 15.86 $\pm$ 2.86 & 16.66 $\pm$ 2.61 & 15.87 $\pm$ 2.84 & 15.04 $\pm$ 2.94 & $\ll$ 0.001\\
ADAS13 & 17.34 $\pm$ 9.03 & 9.08 $\pm$ 4.35 & 16.40 $\pm$ 6.63 & 28.88 $\pm$ 7.84 & $\ll$ 0.001\\
APOE4 & 0.65 $\pm$ 0.68 & 0.34 $\pm$ 0.52 & 0.65 $\pm$ 0.68 & 0.94 $\pm$ 0.70 & $\ll$ 0.001\\
AV45 & 1.24 $\pm$ 0.23 & 1.12 $\pm$ 0.17 & 1.23 $\pm$ 0.23 & 1.43 $\pm$ 0.20 & $\ll$ 0.001\\
ABETA & 927.76 $\pm$ 439.64 & 1161.39 $\pm$ 455.13 & 966.65 $\pm$ 432.87 & 650.40 $\pm$ 294.08 & $\ll$ 0.001\\
FDG & 1.25 $\pm$ 0.09 & 1.30 $\pm$ 0.06 & 1.26 $\pm$ 0.08 & 1.16 $\pm$ 0.11 & $\ll$ 0.001\\
TAU & 299.67 $\pm$ 132.05 & 231.02 $\pm$ 93.54 & 292.18 $\pm$ 128.25 & 368.39 $\pm$ 135.34 & $\ll$ 0.001\\
PTAU & 29.17 $\pm$ 14.62 & 21.20 $\pm$ 9.46 & 28.41 $\pm$ 14.52 & 36.76 $\pm$ 14.22 & $\ll$ 0.001\\
CDRSB & 1.75 $\pm$ 1.67 & 0.04 $\pm$ 0.13 & 1.50 $\pm$ 0.86 & 4.30 $\pm$ 1.60 & $\ll$ 0.001\\
RAVLT\_learning & 4.04 $\pm$ 2.74 & 6.03 $\pm$ 2.36 & 4.09 $\pm$ 2.57 & 1.86 $\pm$ 1.86 & $\ll$ 0.001\\
MMSE & 27.03 $\pm$ 2.60 & 28.99 $\pm$ 1.21 & 27.59 $\pm$ 1.81 & 23.24 $\pm$ 1.97 & $\ll$ 0.001\\
FAQ & 4.43 $\pm$ 6.11 & 0.30 $\pm$ 1.31 & 3.13 $\pm$ 4.07 & 12.73 $\pm$ 6.90 & $\ll$ 0.001\\
\hline
\end{tabular}
}
\label{tab:dataset}
\end{table}

\subsection{AD Subtypes Identification 
}

Here, we focused on detecting the subtypes from the 829 MCI and AD subjects. To demonstrate the effectiveness of similarity learning and graph diffusion on subtype detection, we compared clustering results for the following clustering algorithms on the 829 MCI and AD subjects: (1) $K$-Means; (2) spectral clustering; (3) spectral clustering with graph diffusion; (4) SIMLR; and (5) SIMLR with graph diffusion. Before clustering, all 63 cortical thickness measurements were normalized such that the mean of each measurement is zero and the standard deviation of each measurement is one.
To apply SIMLR for clustering, we utilized the two-step procedure described in (Wang et al 2017) \cite{wang2017visualization} with the SIMLR framework and $K$-Means clustering to cluster the MCI and AD subjects into five subtypes: (1) reduce the data into a 5-dimensional latent space representation using the left singular vectors of the Laplacian matrix computed from the similarity matrix; (2) apply $K$-Means clustering on the latent space representation with a cluster size of 5. 

\begin{figure}[tb]
\centering
\includegraphics[width=0.5\textwidth, clip =true]{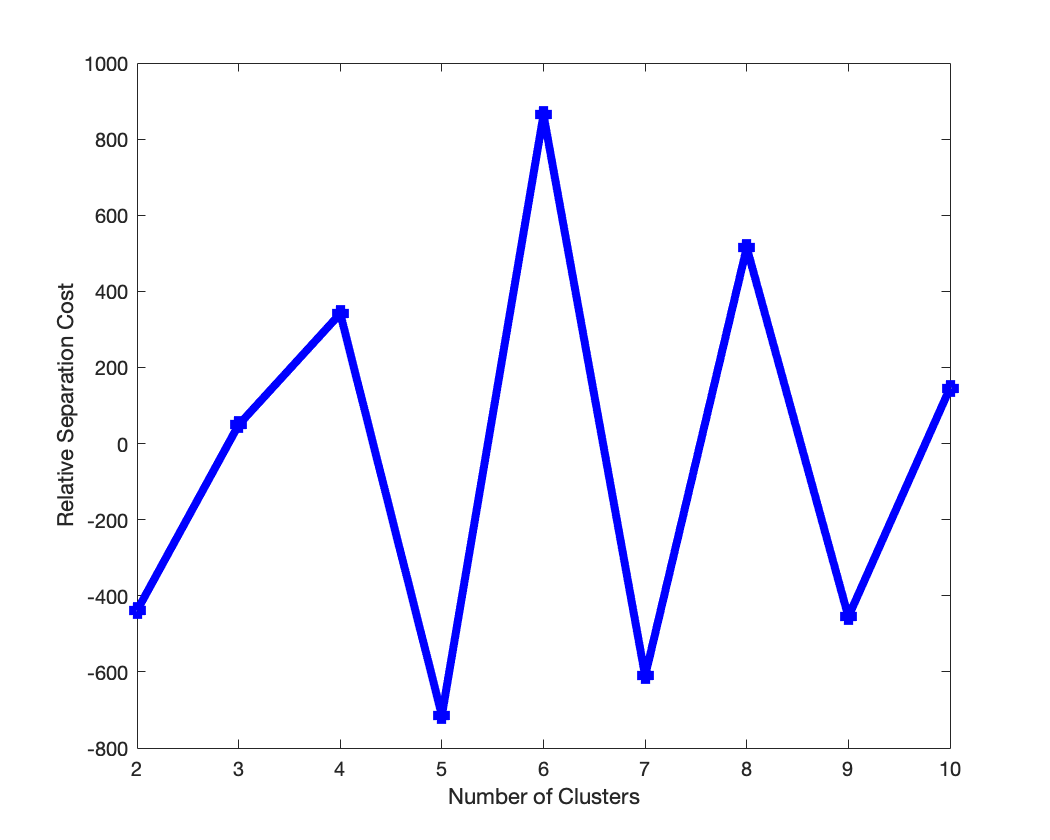}
\caption{Minimizing the separation cost gives the best choice of $K=5$ on the Laplacian of our data.}
\label{fig:opt:K}
\end{figure}

\paragraph{Selection of Cluster Number.} 
As discussed in Section \ref{EXP:diffusion}, we employed a self-tuning approach that selects the cluster number by minimizing the separation cost. The smaller the separation cost is, the more likely the cluster number corresponds to the optimal one. Figure \ref{fig:opt:K} shows the separation cost associated with different choices of the cluster number. We observe that the optimal cluster number $K$ = 5 corresponds to the lowest separation cost.

\begin{figure}[tb]
\centering
\begin{subfigure}[b]{0.475\textwidth}
     \centering
     \includegraphics[width=\textwidth]{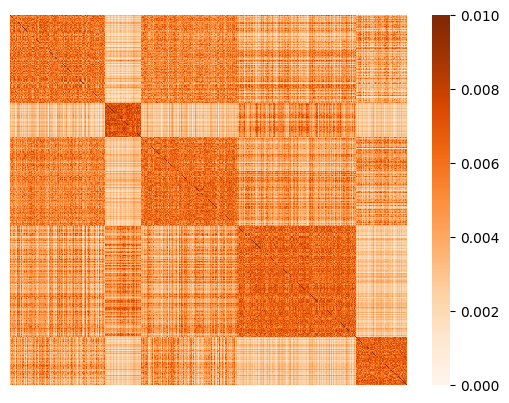}
     \caption{SC}
     \label{fig:simMat_SC}
\end{subfigure}
\hfill
\begin{subfigure}[b]{0.475\textwidth}
 \centering
 \includegraphics[width=\textwidth]{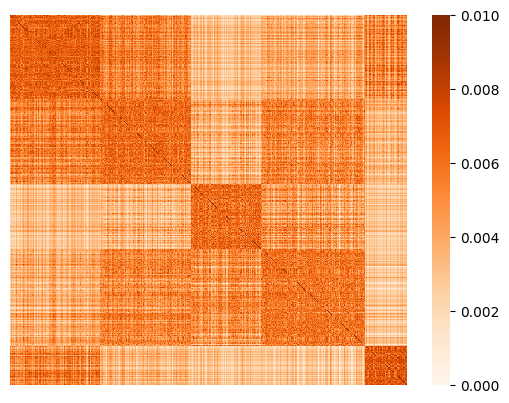}
 \caption{SIMLR}
 \label{fig:simMat_SIMLR}
\end{subfigure}
\vskip\baselineskip
\begin{subfigure}[b]{0.475\textwidth}
     \centering
     \includegraphics[width=\textwidth]{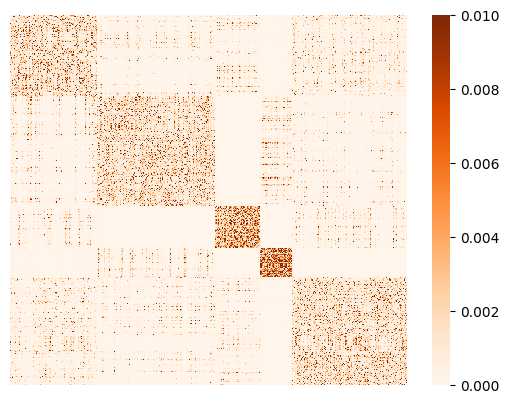}
     \caption{SC + Graph Diffusion}
     \label{fig:simMat_SC_diff}
\end{subfigure}
\hfill
\begin{subfigure}[b]{0.475\textwidth}
 \centering
 \includegraphics[width=\textwidth]{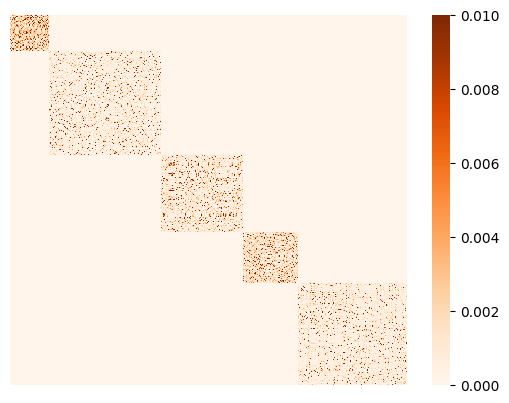}
 \caption{SIMLR + Graph Diffusion}
 \label{fig:simMat_SIMLR_diff}
\end{subfigure}
\caption{Similarity matrices for spectral clustering (SC) and SIMLR}
\label{fig:simMat_all}
\end{figure}

\paragraph{Evaluation of clustering performances.}
As discussed earlier in Section \ref{EXP:diffusion}, here we first showed the effectiveness of the graph diffusion in reducing noise. The similarity matrices used for the four graph-based clustering approaches ($K$-Means excluded) are shown in Figure \ref{fig:simMat_all}. The order of subjects in the similarity matrices were rearranged so that subjects in the same cluster according to the corresponding clustering method were grouped together. All four similarity matrices have a block-diagonal structure. However, the similarity matrices of spectral clustering and SIMLR without graph diffusion suffered from a large amount of noise in the off diagonal entries (Figure \ref{fig:simMat_SC}, \ref{fig:simMat_SIMLR}). After adding the graph diffusion, the noise in the similarity matrices were largely reduced (Figure \ref{fig:simMat_SC_diff}, \ref{fig:simMat_SIMLR_diff}). The noise reduction effect of graph diffusion is especially remarkable under the similarity learning framework. 

\begin{table}[tb]
\caption{Silhouette score comparison. Abbreviation: SC, Spectral Clustering. }
\centering
\begin{tabular}{l c c c c c}
\hline
& \textbf{K-Means} & \textbf{SC} & \textbf{SIMLR} & \textbf{SC + Graph Diffusion} & \textbf{SIMLR + Graph Diffusion} \\
\hline
Silhouette coefficient & 0.1870 & 0.5406 & 0.3577 & 0.6387 & \textbf{0.9788} \\
\hline
\end{tabular}
\label{tab:Silhouette}
\end{table}

Moreover, we compared the performances of different clustering algorithms and adopted the Silhouette coefficient\cite{rousseeuw1987silhouettes} as the metric. The Silhouette coefficient is a widely-used measure that quantifies the quality of clustering results by considering both the cohesion and separation of clusters. Table \ref{tab:Silhouette} showed the Silhouette coefficients for the five clustering methods. For the four similarity-based clustering methods, the Silhouette coefficients were computed on the left singular vectors of the Laplacian matrix obtained from the similarity matrix used for each method. For both spectral clustering and SIMLR, the graph diffusion was able to boost the clustering performance as reflected by the increased Silhouette coefficients. 

\begin{figure}[tb!]
\centering
\begin{subfigure}[b]{0.32\textwidth}
     \centering
     \includegraphics[width=\textwidth]{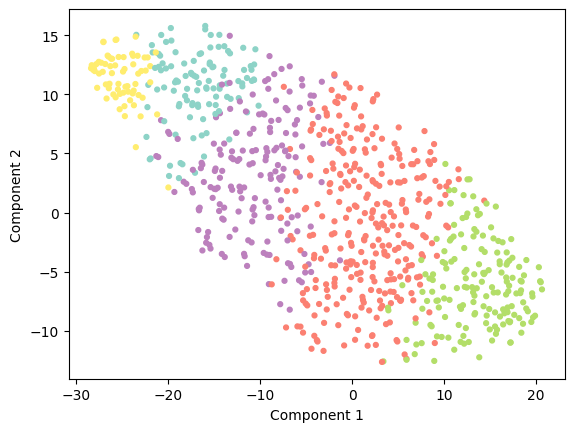}
     \caption{$K$-Means}
     \label{fig:kmeans_vis}
\end{subfigure}
\hfill
\begin{subfigure}[b]{0.32\textwidth}
     \centering
     \includegraphics[width=\textwidth]{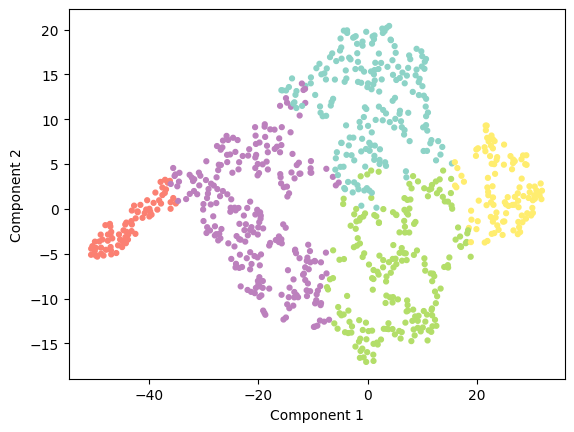}
     \caption{SC}
     \label{fig:SC_vis}
\end{subfigure}
\hfill
\begin{subfigure}[b]{0.32\textwidth}
 \centering
 \includegraphics[width=\textwidth]{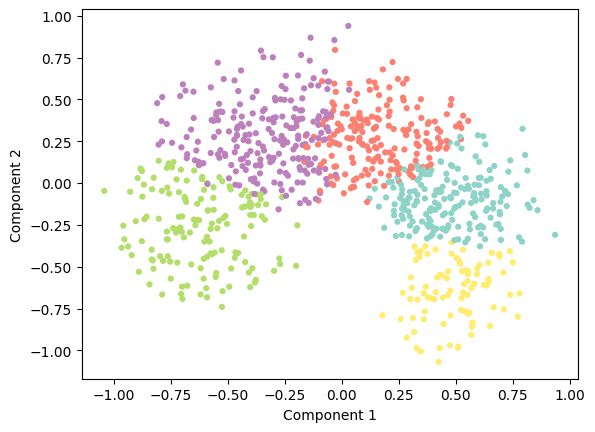}
 \caption{SIMLR }
 \label{fig:SIMLR_vis}
\end{subfigure}
\vskip\baselineskip
\begin{subfigure}[b]{0.32\textwidth}
     \centering
     \includegraphics[width=\textwidth]{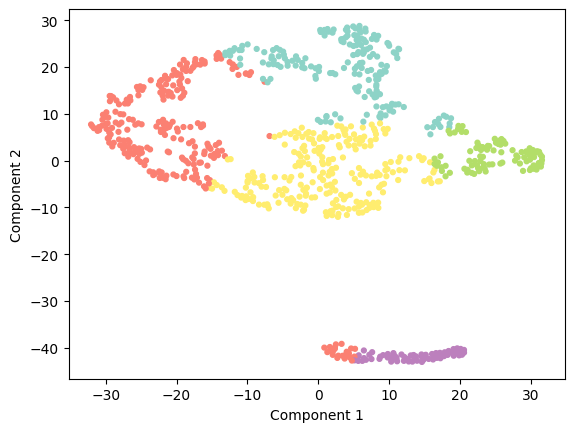}
     \caption{SC + Graph Diffusion}
     \label{fig:SC_diff_vis}
\end{subfigure}
\begin{subfigure}[b]{0.32\textwidth}
 \centering
 \includegraphics[width=\textwidth]{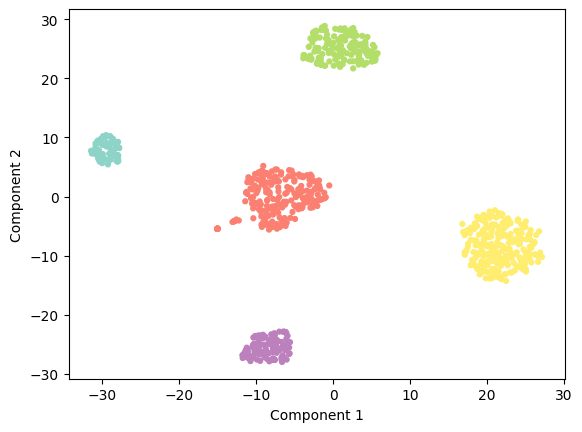}
 \caption{SIMLR + Graph Diffusion}
 \label{fig:SIMLR_diff_vis}
\end{subfigure}
\caption{Two-dimensional visualization with T-SNE for $K$-Means, SC, and SIMLR}
\label{fig:visualization}
\end{figure}

In addition, we applied T-SNE to visualize the two-dimensional embeddings of the clusters; see Figure~\ref{fig:visualization}. For $K$-Means, T-SNE was applied to the normalized data. For the other four graph-based clustering methods, T-SNE was applied to the left singular vector (which is a latent space representation of the normalized data) of the Laplacian matrix. We observed that SIMLR with graph diffusion resulted in the most well-separated visualization for the clusters (Figure \ref{fig:SIMLR_diff_vis}). The following subtype analysis were based on the clustering results obtained from SIMLR with graph diffusion.

\begin{table}[tb]
\caption{Contingency table for the distribution of MCI and AD subjects in the five subtypes. Abbreviations: MCI, mild cognitive impairment; AD, Alzheimer's Disease}
\centering
\begin{tabular}{l c c c c c}
\hline
& \textbf{MCI ($n=631$)} & \textbf{AD ($n=198$)} & \textbf{Total ($n=829$)} \\
\hline
Subtype 1 & 45 & 36 & 81 (9.77\%) \\
Subtype 2 & 162 & 72 & 234 (28.23\%) \\
Subtype 3 & 116 & 55 & 171 (20.63\%) \\
Subtype 4 & 107 & 9 & 116 (13.99\%) \\
Subtype 5 & 201 & 26 & 227 (27.38\%) \\
\hline
\end{tabular}
\label{tab:contingency_table}
\end{table}

\begin{figure}[tb]
\centering
\begin{subfigure}[b]{\textwidth}
     \centering
     \includegraphics[width=\textwidth]{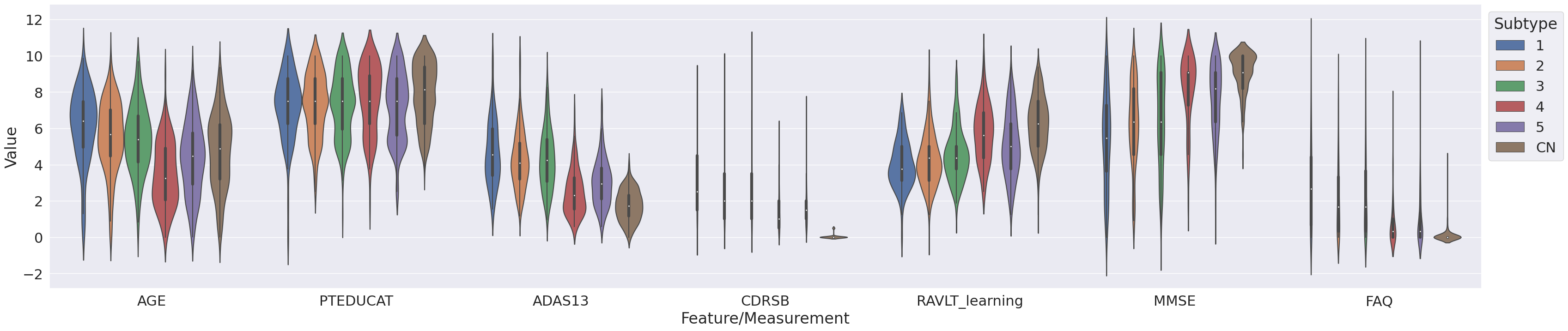}
     \caption{Demographic features and cognitive test scores}
     \label{fig:violin_1}
\end{subfigure}
\vskip\baselineskip
\begin{subfigure}[b]{\textwidth}
     \centering
     \includegraphics[width=\textwidth]{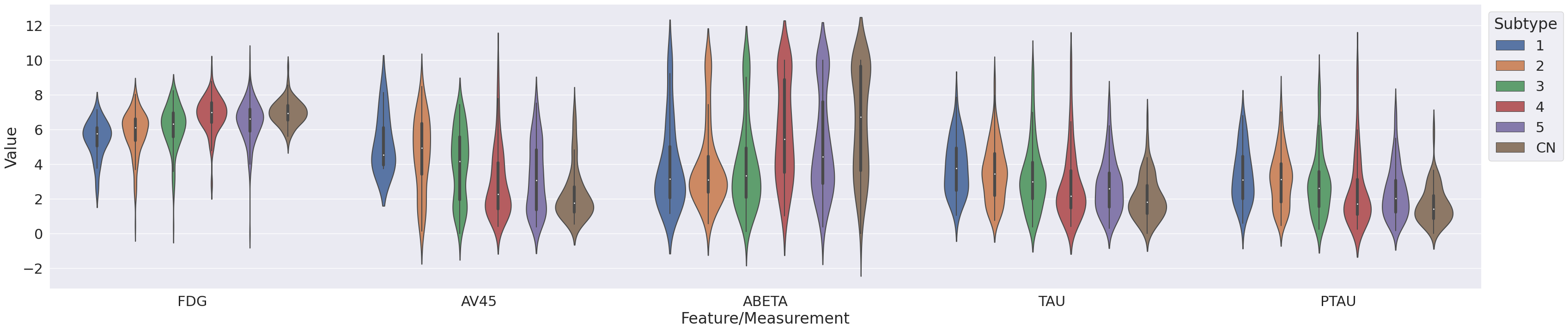}
     \caption{PET imaging and cerebrospinal fluid (CSF) biomarkers}
     \label{fig:violin_2}
\end{subfigure}
\caption{Violin plots of biomarkers, demographical features, and cognitive test scores for each cluster and the control group (CN)}
\label{fig:violin plot}
\end{figure}

\paragraph{Characterization of the clustered subtypes.}
All 829 subjects (631 MCI subjects, 198 AD subjects) were clustered into five subtypes by SIMLR with graph diffusion approach as described above. A contingency table for the distribution of MCI and AD subjects in the five subtypes was shown in Table~\ref{tab:contingency_table}. We performed both the one-way analysis of variance (ANOVA) test and the non-parametric Kruskal-Wallis H-test to compare demographic characteristics, clinical measurements, and biomarkers. Then, we did a Dunn's test for pairwise comparison among the five subtypes and the control group on the same set of features (Table \ref{tab:dunn's test 1}, Table \ref{tab:dunn's test 2}). The five subtypes differed in their demographic features, clinical measurements, and biomarkers (Table \ref{tab:subtype_comp}, Figure \ref{fig:violin_1}, Figure \ref{fig:violin_2}). Subtype 1 was the oldest among all five subtypes. On average, Subtype 1 scored the poorest on all cognitive tests (the highest mean ADAS13, CDRSB, FAQ scores and the lowest mean MMSE and RAVLT scores) among all subtypes, indicating the most severe impairment in cognitive function and daily activities compared to the other four subtypes. Differences from the control group in FDG, TAU, and PTAU results were also the most prominent in Subtype 1, suggesting the most neurofibrillary tangles and the greatest decreased glucose metabolism in the brain among all subtypes. The high AV45 and low CSF ABETA results also indicated the presence of ABETA plaques in the brain. Subtype 2 and Subtype 3 shared similar characteristics as Subtype 1, in which no significant difference in any clinical features and biomarkers was observed between Subtype 1 and 2 and between Subtype 2 and 3. Subtype 3 significantly differed from Subtype 1 with a lower MMSE score and a higher FDG level, suggesting less severe cognitive impairment and less decreased glucose metabolism in the brain. Overall, subjects in Subtypes 1, 2, and 3 were associated with the greatest AD-like characteristics: high ADAS13, CDRSB, FAQ, TAU, and PTAU and low RAVLT score, MMSE, and CSF ABETA. Subjects in Subtype 4 were younger than the other four subtypes. Subtype 4 was also the most similar subtype to the control group with no significant difference in  most of the features. Patients in 5 performed significantly better than the other subtypes in cognitive tests and showed lower levels of CSF ABETA and TAU in the brain. There were no statistically significant difference in years of education among the five subtypes.

\begin{table}[tb]
    \caption{Pairwise comparisons of the five subtypes with the control groups in terms of their clinical and biological characteristics. Dunn's test was performed with Bonferroni correction. Significant p-values (P < 0.05) are in bold. CN: cognitively normal (control)}
    \centering
    \resizebox{\textwidth}{!}{
    \begin{tabular}{lrrrrr}
    \hline
     &  \textbf{Subtype 1 vs CN} &  \textbf{Subtype 2 vs CN} &   \textbf{Subtype 3 vs CN} &   \textbf{Subtype 4 vs CN} &  \textbf{Subtype 5 vs CN} \\
    \hline
    AGE            &          \textbf{5.76e-06} &          \textbf{7.10e-04} &          \textbf{8.07e-02} &          \textbf{1.13e-06} &          6.67e-01 \\
    PTEDUCAT       &          9.61e-02 &          \textbf{4.64e-03} &          \textbf{8.69e-03} &          1.00e+00 &          \textbf{5.80e-04} \\
    ADAS13         &          \textbf{6.50e-38} &          \textbf{1.64e-58} &          \textbf{2.48e-49} &          \textbf{8.12e-04} &          \textbf{2.20e-17} \\
    CDRSB          &          \textbf{3.41e-55} &          \textbf{1.30e-83} &          \textbf{1.03e-71} &          \textbf{1.62e-30} &          \textbf{2.03e-52} \\
    RAVLT\_learning &          \textbf{1.95e-23} &          \textbf{2.52e-34} &          \textbf{1.30e-21} &          1.84e-01 &          \textbf{2.63e-11} \\
    MMSE           &          \textbf{6.97e-32} &          \textbf{2.05e-41} &          \textbf{4.40e-27} &          \textbf{1.07e-03} &          \textbf{3.74e-13} \\
    FAQ            &          \textbf{1.18e-36} &          \textbf{2.87e-47} &          \textbf{1.14e-44} &          \textbf{7.61e-08} &          \textbf{4.43e-17} \\
    FDG            &          \textbf{1.36e-12} &          \textbf{5.44e-14} &          \textbf{3.17e-07} &          1.00e+00 &          \textbf{3.61e-03} \\
    AV45           &          \textbf{3.38e-03} &          \textbf{8.09e-10} &          \textbf{4.18e-08} &          4.82e-01 &          \textbf{1.37e-02} \\
    ABETA          &          \textbf{3.37e-05} &          \textbf{6.45e-09} &          \textbf{2.82e-08} &          1.00e+00 &          6.23e-02 \\
    TAU            &          \textbf{3.60e-08} &          \textbf{5.23e-10} &          \textbf{6.89e-06} &          2.89e-01 &          \textbf{2.00e-02} \\
    PTAU           &          \textbf{4.68e-08} &          \textbf{1.40e-11} &          \textbf{8.93e-07} &          2.20e-01 &          \textbf{6.34e-03} \\
    \hline
    \end{tabular}}

\label{tab:dunn's test 1}
\end{table}

\begin{table}[tb]
\caption{Pairwise comparisons among the five subtypes in terms of their clinical and biological characteristics. Dunn's test was performed with Bonferroni correction. Significant p-values (P < 0.05) are in bold. }
    \centering
    \resizebox{\textwidth}{!}{
    \begin{tabular}{lrrrrrrrrrr}
    \hline
    \textbf{Subtypes} &  \textbf{1 vs 2} &  \textbf{1 vs 3} &  \textbf{1 vs 4} &  \textbf{1 vs 5} &  \textbf{2 vs 3} &  \textbf{2 vs 4} &  \textbf{2 vs 5} &  \textbf{3 vs 4} &  \textbf{3 vs 5} &  \textbf{4 vs 5} \\
    \hline
    AGE & 0.48 & 0.07 & \textbf{4.90e-18} & \textbf{3.65e-10} & 1.00 & \textbf{3.81e-18} & \textbf{4.30e-09} & \textbf{3.38e-13} & \textbf{2.46e-05} & \textbf{2.22e-03} \\
    PTEDUCAT & 1.00 & 1.00 & 1.00 & 1.00 & 1.00 & 9.33e-01 & 1.00 & 9.611e-01 & 1.00e+00 & 0.31 \\
    ADAS13 & 1.00 & 1.00 & \textbf{6.23e-17} & \textbf{2.75e-10} & 1.00 & \textbf{3.52e-21} & \textbf{2.19e-13} & \textbf{2.20e-18} & \textbf{6.79e-11} & \textbf{0.01} \\
    CDRSB & 1.00 & 1.00 & \textbf{1.12e-05} & \textbf{8.21e-05} & 1.00 & \textbf{9.32e-05} & \textbf{5.41e-04} & \textbf{2.83e-04} & \textbf{2.05e-03} & 1.00 \\
    RAVLT\_learning & 1.00 & 0.24 & \textbf{4.28e-12} & \textbf{4.38e-06} & 1.00 & \textbf{2.05e-14} & \textbf{5.29e-07} & \textbf{1.74e-08} & \textbf{1.23e-02} & 0.01 \\
    MMSE & 0.66 & \textbf{0.02} & \textbf{2.08e-13} & \textbf{1.39e-09} & 1.00 & \textbf{7.66e-13} & \textbf{7.78e-09} & \textbf{1.81e-07} & \textbf{7.94e-04} & 0.23 \\
    FAQ & 0.38 & 1.00 & \textbf{3.49e-11} & \textbf{7.99e-10} & 1.00 & \textbf{2.10e-09} & \textbf{2.41e-08} & \textbf{4.55e-10} & \textbf{6.13e-09} & 1.00 \\
    FDG & 0.52 & \textbf{2.38e-03} & \textbf{1.29e-11} & \textbf{1.42e-06} & 0.15 & \textbf{7.28e-13} & \textbf{7.08e-06} & \textbf{3.34e-06} & 2.62e-01 & \textbf{0.02} \\
    AV45 & 1.00 & 1.00 & 5.79e-02 & 0.19 & 1.00 & \textbf{2.48e-05} & \textbf{1.17e-03} & \textbf{2.39e-03} & 9.81e-02 & 1.00 \\
    ABETA & 1.00 & 1.00 & \textbf{4.45e-04} & 0.06 & 1.00 & \textbf{3.44e-07} & \textbf{9.44e-04} & \textbf{1.38e-06} & \textbf{2.98e-03} & 0.59 \\
    TAU & 1.00 & 0.23 & \textbf{3.78e-04} & \textbf{1.18e-03} & 1.00 & \textbf{2.76e-04} & \textbf{7.52e-04} & 0.13 & 0.41 & 1.00 \\
    PTAU & 1.00 & 0.54 & \textbf{6.53e-04} & \textbf{3.84e-03} & 0.70 & \textbf{3.77e-05} & \textbf{2.63e-04} & 0.05 & 0.34 & 1.00 \\
    \hline
    \end{tabular}}
\label{tab:dunn's test 2}
\end{table}

\begin{table}[tb]
\centering
\caption{Comparison of the five subtypes in terms of their clinical and biological characteristics. Both a parametric one-way ANOVA test and a non-parametric Kruskal-Wallis test were performed.}
\begin{tabular}{lrr}
\hline
 & \textbf{P (one-way ANOVA)} & \textbf{P (Kruskal–Wallis test)} \\
\hline
AGE &  3.19e-29 &  8.76e-28 \\
PTEDUCAT &  2.64e-01 &  2.22e-01 \\
ADAS13 &  2.70e-40 &  1.01e-38 \\
CDRSB &  3.89e-15 &  1.07e-14 \\
RAVLT\_learning &  2.39e-21 &  8.69e-20 \\
MMSE &  1.25e-24 &  5.94e-23 \\
FAQ &  5.58e-20 &  8.98e-24 \\
FDG &  9.29e-15 &  2.02e-17 \\
AV45 &  2.39e-07 &  2.99e-07 \\
ABETA &  1.07e-09 &  3.62e-10 \\
TAU &  4.52e-05 &  4.03e-07 \\
PTAU &  8.01e-05 &  1.91e-07 \\
\hline
\end{tabular}
\label{tab:subtype_comp}
\end{table}

\subsection{Genetic Association Analysis of Identified AD Subtypes}

\begin{figure}[tb]
\centering
\includegraphics[width=\textwidth]{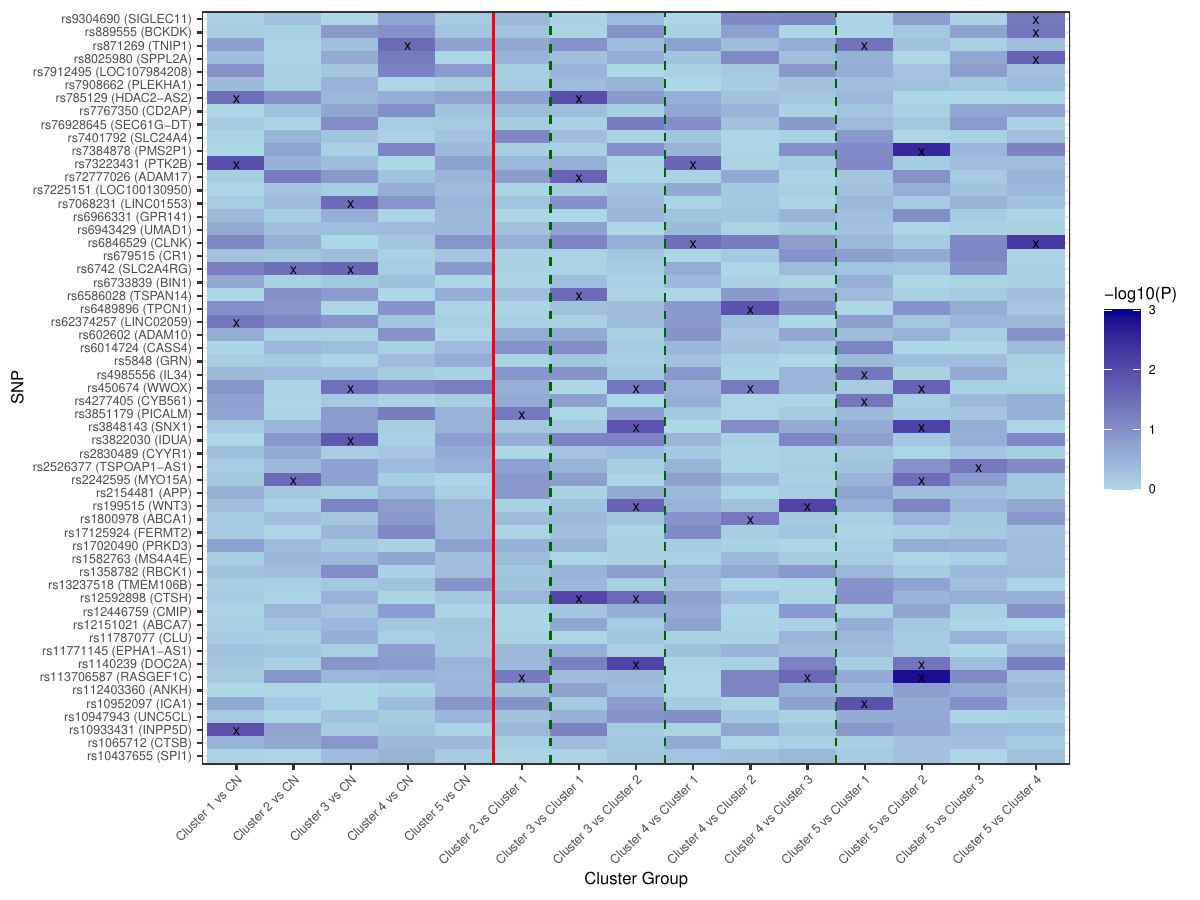}
\caption{Targeted genetic association analysis. \iffalse AD is complex and neuropathologically multifaceted \cite{schafer2013personal,wen2022genetic,bao2021identifying,bao2021identifying2,bao2023integrative}. Studying the genetics of AD subtypes may offer insights into the genetic etiological mechanisms for different neuropathological features of AD. In this study, we used SIMLR to cluster subjects into five AD subtypes based on their regional brain thickness.\fi We conducted the targeted genetic association analysis and compared the resulting AD subtypes with cognitively normal subjects (left of the red line) and compared different subtypes of AD (right of the red line). Univariate logistic regressions were performed and the $-\log10(p$-$values)$ were plotted. We marked all the significant associations by "x" ($p$-$value<0.05$). All the SNPs were annotated by their nearest gene using ANNOVAR \cite{wang2010annovar}.}
\label{fig:targetedG}
\end{figure}

To evaluate our clustering findings, we performed targeted genetic association analysis. The selection of targeted genetic variants was based on the large-scale genome-wide association studies (GWAS) of Alzheimer's disease (AD) and closely related phenotypes to identify relevant single nucleotide polymorphisms (SNPs) (See Table 1 in \url{https://adsp.niagads.org/gvc-top-hits-list/}). These SNPs were carefully curated and authenticated by the Gene Verification Committee of the Alzheimer's Disease Sequencing Project Consortia. The detailed methodology for prioritizing these SNPs can be found at \url{https://adsp.niagads.org/wp-content/uploads/2022/05/SingleVariantTest-V42.pdf}, ensuring strong genetic associations with AD. This rigorous genetic evidence supports the credibility and importance of our targeted SNPs in the context of studying AD and AD related dementias (ADRD).

In our study, we used a univariate logistic regression to identify genetic variants that are significantly associated with five AD subtypes compared to cognitively normal subjects. To assess the genetic markers that differentiate different AD subtypes, we also conducted targeted genetic association analyses between different groups of AD subtypes. To control for potential confounders, we adjusted the association tests using age, sex, and population structure information quantified by first 10 principal components derived from imputed genotyping data. Figure~\ref{fig:targetedG} shows the genetic association results. Our analysis identified multiple genetic markers significantly associated with each of the five AD subtypes compared to the cognitively normal group, indicating successful differentiation between AD subtypes based on genetic markers. Furthermore, the significant genetic markers identified in our analysis also suggest that patients from different AD subtypes are genetically distinct. Overall, the targeted genetic analysis characterizes our clustering findings and provides potential genetic etiological mechanisms for different AD subtypes.

\section{DISCUSSION}

As the most prevalent neurodegenerative disease, currently a major challenge in AD is to detect its subtypes which are known to have different rates of cognitive declines and disease pathologies. In this study, we presented an approach that discovers AD subtypes using graph-based similarity learning on brain cortical thickness measures obtained from MRI data. The method performs graph learning and community inference at the same time by utilizing a combination of the SIMLR framework \cite{wang2017visualization} and  graph diffusion  \cite{yang2010modeling} to jointly estimate the pairwise similarities and subpopulation clusters of AD and MCI subjects. This allows the method to create observably better separation of the clusters compared to conventional AD subtyping methods, such as $K$-Means or Spectral Clustering, that only use predefined similarity metrics to do clustering. We have further inspected the mechanism of the SIMLR method and analyzed which step improves its performance. We have concluded that the main contribution to distinguishable clusters came from graph diffusion, a technique of noise reduction. We noticed that graph diffusion is also capable of improving the separability of the clusters for other methods, such as Spectral Clustering. Our analysis revealed five distinct subtypes in AD and MCI subjects with different clinical and biological characteristics. 
Below we discuss a few relevant topics regarding the methods and results of our study.

\paragraph{Potential Genetic Targets for AD Research.}
To further verify the result, we conducted a genetic association analysis and discovered the corresponding genetic basis for the five subtypes. Our findings may provide new insights into the subtypes of AD and MCI. As shown in the above Section~\ref{Results}, Subtype 1 showed the most prominent differences in biomarkers and cognitive test results from the control group, and Subtype 1 was the oldest among all subtypes. The genetic basis analysis showed that Subtype 1 had a significantly different association with INPP5D compared to the control group, indicating that mutations in INPP5D are associated with an increased risk of AD. This finding is consistent with previous studies that have identified INPP5D as a potential therapeutic target for late-onset AD (LOAD) \cite{lambert2013meta,jing2016inpp5d} through its expression in plaque-associated microglia \cite{tsai2021inpp5d} which is positively correlated with amyloid beta plaque density. Our result also suggests that INPP5D may be a potential target for future AD research and therapeutics.
Of note, in this analysis, we employed a generous p threshold without correcting for multiple comparison. As we know, all the AD genetic signals except APOE e4 allele have small effect sizes. Given the modest sample size of our data, our genetic association study is under-powered. By adopting a relaxed threshold, we aim to ensure a more flexible exploration of the data and a greater ability to suggest potentially meaningful genetic signals. It warrants further investigation to replicate and confirm these identified signals in larger and independent cohorts to ensure statistical rigor.

\paragraph{Enhancement on Kernels.}
Incorporating known labels, such as AD, control, and MCI, can be valuable in identifying important graph edges or kernels to refine results. One approach could be to introduce a group Lasso term with an adjustable regularization parameter $\rho_4$ to the SIMLR objective function (\ref{eqn:obj:simlr}), inspired by the works of (Yuan and Lin 2006) \cite{yuan2006model} and (Danaher et al. 2014) \cite{danaher2014joint}. This term promotes similarity in magnitude among the entries of the graph matrix corresponding to nodes with shared labels, effectively shrinking them together. By adjusting the regularization level, denoted as $\rho_4$, we can control the extent of shrinkage and introduce sparsity at the community level. Higher values of $\rho_4$ result in increased penalty and greater sparsity, allowing for a more refined estimation of the graph that reflects the underlying grouping structure of nodes based on their labels. This incorporation of label information enhances the accuracy of the results by leveraging the knowledge about the community structure within the data. Further investigation and exploration are needed to accurately estimate this structure. 

\paragraph{Generalizability.}
The performance of the ``SIMLR + graph diffusion'' approach in clustering analysis is influenced by various factors, including the specific analysis conducted, characteristics of the dataset used, and the underlying assumptions of the algorithm. It is important to understand that no single clustering algorithm can be considered universally optimal for all datasets, as each algorithm operates based on its own set of assumptions. While graph diffusion, as implemented in SIMLR, tends to yield cleaner separation of clusters compared to methods like spectral clustering, its effectiveness may still vary depending on the specific dataset being analyzed. Our approach aims to achieve accurate clustering through a high-level diffusion process. However, we acknowledge the potential risk of overfitting when attempting to completely separate clusters using the ``SIMLR + graph diffusion'' approach.

To demonstrate the generalizability of our method, we performed a comparative analysis with spectral clustering, a well-established clustering technique. Our findings revealed that the ``SIMLR + graph diffusion'' method outperformed spectral clustering in identifying distinct clusters among AD and MCI patients. Additionally, we examined the biological and clinical characteristics, as well as the genetic basis, across different clusters identified by our method. The clusters obtained through ``SIMLR + graph diffusion'' exhibited meaningful patterns that differentiated them from each other and from the control group.

To further address concerns regarding overfitting, future studies should focus on validating and applying the clusters identified by the ``SIMLR + graph diffusion'' approach to datasets beyond the current one. Replicating the findings using additional datasets would provide further support for the reliability of the clustering analysis. Furthermore, it would be valuable to consider potential sources of bias and confounding factors to obtain a comprehensive understanding of the clustering results. By addressing these aspects, we can enhance our understanding of the validity and generalizability of the clustering results obtained with the ``SIMLR + graph diffusion'' method.

\paragraph{AD Subtyping.} While the usage of graph-based similarity learning approach has gained considerable success in other domains, little work has been done previously for AD subtyping leveraging this approach.  Here, our study adapted the SIMLR framework and introduced it to the AD subtyping task for the first time, which may serve as a pioneering study for future research. For example, in the current setting, we only focused on the cortical thickness measurements from brain MRI; but future research could incorporate other features from imaging and other modalities to provide more comprehensive perspectives to improve the pairwise similarity among subjects. Furthermore, the learned subjects similarity matrix can also be used to prioritize cortical thickness or other putative features to identify which features, across all subject, correlate the best with the learned pairwise similarity. This would help find important markers that are indicative of different AD subtypes. In addition, SIMLR's scalability strategy would allow for exploration of much larger cohorts using k-nearest neighbor approximation of the full pairwise similarity, when more data become available in the future.

\section*{Acknowledgments}

This work was supported in part by National Institutes of Health grants U01 AG068057, U01 AG066833, R01 AG071470, RF1 AG063481 and RF1 AG068191; and National Science Foundation grant IIS 1837964.

Data collection and sharing for this project was funded by the Alzheimer's Disease Neuroimaging Initiative (ADNI) (National Institutes of Health Grant U01 AG024904) and DOD ADNI (Department of Defense award number W81XWH-12-2-0012). ADNI is funded by the National Institute on Aging, the National Institute of Biomedical Imaging and Bioengineering, and through generous contributions from the following: AbbVie, Alzheimer's Association; Alzheimer's Drug Discovery Foundation; Araclon Biotech; BioClinica, Inc.; Biogen; Bristol-Myers Squibb Company; CereSpir, Inc.; Cogstate; Eisai Inc.; Elan Pharmaceuticals, Inc.; Eli Lilly and Company; EuroImmun; F. Hoffmann-La Roche Ltd and its affiliated company Genentech, Inc.; Fujirebio; GE Healthcare; IXICO Ltd.; Janssen Alzheimer Immunotherapy Research \& Development, LLC.; Johnson \& Johnson Pharmaceutical Research \& Development LLC.; Lumosity; Lundbeck; Merck \& Co., Inc.; Meso Scale Diagnostics, LLC.; NeuroRx Research; Neurotrack Technologies; Novartis Pharmaceuticals Corporation; Pfizer Inc.; Piramal Imaging; Servier; Takeda Pharmaceutical Company; and Transition Therapeutics. The Canadian Institutes of Health Research is providing funds to support ADNI clinical sites in Canada. Private sector contributions are facilitated by the Foundation for the National Institutes of Health (www.fnih.org). The grantee organization is the Northern California Institute for Research and Education, and the study is coordinated by the Alzheimer's Therapeutic Research Institute at the University of Southern California. ADNI data are disseminated by the Laboratory for Neuro Imaging at the University of Southern California. 

This work uses the TADPOLE data sets https://tadpole.grand-challenge.org constructed by the EuroPOND consortium http://europond.eu funded by the European Union's Horizon 2020 research and innovation programme under grant agreement No 666992.

\section*{Declaration of interests}

The authors declare no competing interests.

\section*{Author contributions section}

Conceptualization, T.W., S.Y., D.A.T. and L.S.;
Methodology, T.W., S.Y., and D.A.T.; 
Investigation, T.W., S.Y., D.A.T., J.B., J.X., P.O., J.B.W., and L.S.; 
Writing – Original Draft, T.W., S.Y., and D.A.T.; 
Writing – Review \& Editing, T.W., S.Y., D.A.T., J.B., J.X., P.O., J.B.W., Q.L., and L.S.; 
Funding Acquisition, L.S. and Q.L.; 
Supervision, L.S.

\bibliographystyle{vancouver}

\bibliography{ref_simlr.bib}
\end{document}